\documentclass{sigchi-ext}
\usepackage[T1]{fontenc}
\usepackage{textcomp}
\usepackage[scaled=.92]{helvet} % for proper fonts
\usepackage{graphicx} % for EPS use the graphics package instead
\usepackage{balance}  % for useful for balancing the last columns
\usepackage{booktabs} % for pretty table rules
\usepackage{ccicons}  % for Creative Commons citation icons
\usepackage{ragged2e} % for tighter hyphenation
\usepackage{multirow}
\def\plaintitle{Learning with Collaborative Neural Network Group by Reflection} 
\def\emptyauthor{}
\def\plainkeywords{Deep learning; neural networks; biology-related information processing; interpretability of deep learning}

\title{Learning with Collaborative Neural Network Group by Reflection}

\numberofauthors{3}
\author{%
  \alignauthor{%
  \textbf{Liyao Gao}\\
    \affaddr{Purdue University}\\
    \affaddr{West Lafayette, IN 47907, USA}\\
    \email{gao463@purdue.edu}}\\
    \vfil
  \alignauthor{%
  \textbf{Zehua Cheng}\\
  \affaddr{Fuzhou University}\\
  \affaddr{Fuzhou, China}\\
  \email{limber@snowcloud.ai}}\\
    }
\definecolor{linkColor}{RGB}{6,125,233}
\hypersetup{%
  pdftitle={\plaintitle},
  pdfauthor={\emptyauthor},
  pdfkeywords={\plainkeywords},
  bookmarksnumbered,
  pdfstartview={FitH},
  colorlinks,
  citecolor=black,
  filecolor=black,
  linkcolor=black,
  urlcolor=linkColor,
  breaklinks=true,
}
\begin{document}
%% For the camera ready, use the commands provided by the ACM in the Permission Release Form.
% \CopyrightYear{2018}
% \setcopyright{rightsretained}
% \conferenceinfo{WOODSTOCK}{'97 El Paso, Texas USA}
% \isbn{0-12345-67-8/90/01}
% \doi{http://dx.doi.org/10.1145/2858036.2858119}
%% Then override the default copyright message with the \acmcopyright command.
% \copyrightinfo{\acmcopyright}
\maketitle

% Uncomment to disable hyphenation (not recommended)
% https://twitter.com/anjirokhan/status/546046683331973120
% \RaggedRight{} 

% Do not change the page size or page settings.
\begin{abstract}
  For the present engineering of neural systems, the preparing of extensive scale learning undertakings generally not just requires a huge neural system with a mind boggling preparing process yet additionally troublesome discover a clarification for genuine applications. In this paper, we might want to present the Collaborative Neural Network Group (CNNG). CNNG is a progression of neural systems that work cooperatively to deal with various errands independently in a similar learning framework. It is advanced from a solitary neural system by reflection. Along these lines, in light of various circumstances removed by the calculation, the CNNG can perform diverse techniques when handling the information. The examples of chose methodology can be seen by human to make profound adapting more reasonable. In our execution, the CNNG is joined by a few moderately little neural systems. We give a progression of examinations to assess the execution of CNNG contrasted with other learning strategies. The CNNG is able to get a higher accuracy with a much lower training cost. We can reduce the error rate by 74.5\% and reached the accuracy of 99.45\% in MNIST with three feedforward networks (4 layers) in one training epoch.
\end{abstract}


\category{Computing methodologies}{Learning from implicit feedback}{}

\section{Introduction}
Researchers have focused a lot on neural networks. In recent years, a series of deep neural networks has been introduced. Many deep neural networks can reach a satisfying performance in many tasks, including image recognition, speech recognition and machine translation~\cite{lecun2015deep}. Currently, a common belief is the building of a deeper neural network is able to solve a very larger learning task, such as the using of VGG~\cite{simonyan2014very} and ResNet~\cite{he2016deep}. It may have a satisfying performance. However, we believe that only making the neural network deeper might be an optimal choice to solve a large-scale task. Specifically, there can be two disadvantages of using a single deep neural network to process a large-scale learning task.
For the first point, using a single neural network to learn a large-scale learning task might require more complex technique in optimization. The current performance of the neural network seems to be difficult to improve the performance for a further step. Take handwritten recognition task here as an example. The typical error cases that result in the error can be really ambiguous. It is difficult for the neural network to escape from the local minima. The using of a single neural network to solve the large task may increase the difficulty of training and optimization. Obviously, a deeper network might make the problem even worse. Secondly, the expansion in size of a deep neural network might not be efficient. A large-scale task can usually be separated into several subtasks. For example, EMNIST\cite{cohen2017emnist} is a datasets of handwritten recognition tasks combined with digits and letters. Intuitively, the EMNIST task can be separated into digit recognition task and letters recognition task. The optimal overall error might require completely different parameters compared to only focusing on the error in digits. If we try to use a single deep neural network to solve the problem, the combination of these two situations might require a network that much larger in size.

In this paper, we would like to introduce the Collaborative Neural Network Group (CNNG) by reflection. Collaborative Neural Network Group is an architecture that combined with a series of neural networks. Reflection is the learning algorithm that is able to generate the CNNG from a single neural network. It is originated from the learning strategy of humans~\cite{boud2013reflection}. In a learning task, human will perform a reflection that reconsiders the problem and analyze their mistakes. As a similar approach in a neural network way, a general neural network will be initially trained for a learning task. Then, the error cases of the general network will be collected. The algorithm will classify the error cases into different clusters and initialize a corresponding number of neural networks to be trained by the error clusters. The networks that focus on the different error cases will become the specialist neural network. For the last step, a task classifier will be trained based on the error clusters to determine which network to use for an incoming data. This is the way how a single neural network will be evolved into the CNNG by reflection. The networks in CNNG are viewed to be different strategies to use when processing the tasks.

From our perspective, the CNNG by reflection can be a good method to use for a large-scale learning task. Firstly, it is able to lower the difficulty of optimization. In the CNNG, the ambiguous cases are separated into different clusters and will be processed by different networks. Therefore, it would decrease the difficulty of optimization. Also, after reflection, the specialist networks will be used on their specified tasks. It is able to have a better performance compared to the general network on those tasks. In this way, for every time that the task classifier is making a correct assignment, the CNNG will have a higher probability to predict the output correctly. The using of the specialist network will not only help the CNNG to use a network with higher accuracy on the task, but also remove the cases that originally is ambiguous for the general network. Therefore, the accuracy can be greatly improvement. From this point, we believe that CNNG by reflection can be a better method to solve the problem of large-scale learning tasks. We view this as an important method to solve large-scale learning tasks with a higher efficiency and accuracy.

A series of experiments has been provided in this paper. We compared the performance of CNNG by reflection with other typical learning methods. We tested the result thoroughly on two image datasets, MNIST and EMNIST. Our experiment reached a satisfying result and our model can largely lower the loss. Specifically for the EMNIST, the CNNG after reflection is able to reduce the error rate by 46.8\%. We reached the accuracy of 90.88\% with three simple feedforward networks (4 layers) with one training epoch. For MNIST, we are able to reach the accuracy of 99.45\% with three simple

feedforward networks (3 layers) with one training epoch. It lowers the error rate by 74.5\%. Details can be found in the Evaluation part.

Our main contribution in this paper is the introduction of the Collaborative Neural Network Group by reflection.

It is able to process the large-scale learning task with a high accuracy with a much lower training cost. The specific design of the reflection algorithm have been discussed. The choice of task classifier and number of specialist networks has been explained. Detailed evaluation has been provided with the existing popular methods in learning. Our experiment result shows its superiority in both efficiency and accuracy.

\section{CNNG Architecture}

CNNG is a group of neural networks that will be used to process the input data. It is evolved from a single neural network by the reflection algorithm. The benefit of CNNG is the collaboration of several small networks which are specified on different special tasks individually can help to largely improve the accuracy. It requires a much lower training cost at the same time. The architecture of the CNNG is combined with task classifier, general neural network and specialist neural networks.

Task Classifier: The task classifier is used for the CNNG to decide which is the best network to use when predicting the label of the input data. In our approach, the task classifier is a decision tree. The task classifier allows the neural networks to collaboratively work together as a group.

General neural network: The general neural network is the network that is initially trained by all the training data. This will serve as a general situation when handling the input data. A reflection will be performed based on the general network. The error cases will be used to train the specialist neural networks.

Specialist neural network: The specialist neural network is the neural network that will focus on different subtasks. It will be trained from the error that produced by the general network. It is viewed as the specialist in the system that will process the corner cases for general network.

For the prediction process of CNNG, the task classifier will first determine the best network to use for an input data. Then, the best network will take charge of the input data and predict the label.

\section{Reflection}

The reflection algorithm here is motivated by learning strategy of humans. Humans frequently perform a reflection when they are approaching their extreme in performance in a learning task. Reflection is an important process in learning~\cite{boud2013reflection}. Normally, as we stated before, human will try to analyze the problem, divide them into different cases, and try to apply different methods to solve them separately. Based on this idea, we designed the concept of reflection on Collaborative Neural Network Group. Basically, the general neural network that will be trained initially to handle the general situation. This is served as the general training process. For the cases that still have a high error for the general network, it will be divided into different clusters. Then, a series of new neural networks will be initialized and trained with the input with a high error for the general neural network. This is served as the second training process. For the last step, the task classifier will be trained with inputs and labels of the network id. The task classifier is going to decide which network to use for input. We use the K-Means method here to decompose the error cases.

The task classifier plays an important role in reflection. It is served as a task assigner or a strategy decider in CNNG which will largely affect the performance. In the process of reflection, the input data are separated by features that determined by the error cases. The choice of task classifier should pair with the method we use in the reflection that is able to understand the patterns for the best. In our implementation, K-Means is used in splitting the error cases while reflecting. Because of the using of K-Means, the pattern of the label can be viewed as a group of constraints by numbers for x.

In this case, the decision tree method could possibly be the best classifier here. The decision tree uses a threshold to classify the data into different groups and this is the pattern that we would like the task classifier to learn. Based on our experiments, the decision tree method does have the best performance compared to other kinds of classical classifiers. Details can be found in the evaluation part.

\section{Experiments and Evaluation}
We applied the CNNG by reflection into different situations. We compared the general performance of CNNG with reflection compared to other methods. In this paper, we focused on the evaluation on the image tasks. The evaluation of task classifier and the networks in CNNG individually can provide an insight to show where does the increase in performance come from.

The SingleNN that we use in the following experiments is a simple feedforward network. For the MNIST, the initial network contains three layers. For the EMNIST, the initial network contains four layers.

\section{Overall performance of CNNG to other methods}
In this section, we provided an overall performance of CNNG and its comparison to other classic methods. The CNN here is combined with two convolutional layers and four linear layers. Adaboost is a statistical method of boosting neural networks~\cite{schwenk2000boosting}. Four classifiers is used here. CNNE~\cite{islam2003constructive} is another incremental learning method which sequentially initiate a new network on the error. We evaluate the performance of CNNG by reflection with existing methods. For the CNNG here, it is combined with three simple feedforward networks. All of the following methods are trained with one epoch. We choose the type balanced in EMNIST, which contains 131000 images for 62 different outputs. The Adaboost method is using four classifiers.

\begin{table}[!ht]
  \centering
  \begin{tabular}{l|lll}
  &CNNG(*) & SingleNN&CNN\\\hline
  MNIST&99.45\%&97.84\%&96.32\%\\
  EMNIST&\textbf{90.88}\%&82.32\%&83.5\%\\
  \end{tabular}
\end{table}
Firstly, as we can see, the CNNG is having the best performance among other methods with low training effort. The test accuracy that it is able to reach is satisfying for both 99.45\% in MNIST and 90.88\% in EMNIST. It lowers the error rate by 48.4\% in EMNIST and 74.5\% for MNIST. Especially for the EMNIST, it improves largely compared to 78.02\% accuracy on OPIUM classifier~\cite{cohen2017emnist}. It seems to be the best performance so far. Another point here is that under the condition of training with only one epoch, the CNNG is having a better performance to CNN. Based on this experiment, we can see that CNNG by reflection is able to reach the highest performance than other methods by simple neural networks with low training efforts.

This experiment aims to explain a detailed observation behind the improved performance in CNNG and provide evaluation of the reflection algorithm. The following shows the result of the networks in CNNG individually. It evaluates the accuracy on specified task is resulted by cross validation and the times that each network is used by CNNG in prediction. We also test the accuracy of all three networks of its overall performance on the whole task. We conducted this experiment using EMNIST.

\begin{table}[!ht]
  \centering
  \begin{tabular}{l|lll}
  &\multirow{2}{*}{}CNNG & Overall&Specific task \\
  &Used&Accuracy&Accuracy\\\hline
  GenNet&72.5\%&82.32\%&82.32\%\\
  SpecNet1&12.4\%&15.32\%&95.33\%\\
  SpecNet2&14.9\%&10.20\%&94.32\%\\
  \end{tabular}
\end{table}

According to this result, there are two points that we can summarize. Firstly, the reflection is successfully decomposing a large learning task into easier subtasks. More specifically, the specialist network is able to have a better performance in the specified tasks and the specified tasks have a lower difficulty in training. As we can see in the table, for the two specialist networks, they have a very low overall accuracy in overall, which are both around 10\% to 20\%. However, in their specified task, they have a much higher accuracy compared to general network. An important point here is the specialist networks are only trained with the error cases, which is only 3\% of the data in all. It can infer that the specified tasks are easier to be learned. Secondly, we can see that the overall performance of the general network that trained by all the data can reach the accuracy of 82.32\%. However, the CNNG after reflection is able to reach 90.88\%. The specialist networks are able to have a higher performance in their specific tasks. Therefore, when the task classifier is assigning the task into the specialist networks, it is going to have a higher probability to predict the correct output compared to use the general network. This experiment explains the improved performance of CNNG and provides strong evidence that the reflection algorithm is successfully decomposing the large tasks into easier subtasks.

\section{Discussion}
In our implementation of reflection, it is unable for the algorithm to dynamically decide k, the number of specialist network to use. As we stated before, the choice of k can largely affect the performance of CNNG. However, it should be able to determine by analyzing the error cases. The desired k here should try to minimize the difficulty of the training of task classifier and specialist neural networks. A kernel problem here is the estimation of the difficulty for a training set to be learned by a neural network.

For future works, Firstly, a more general reflection algorithm and cohesive learning theory should be introduced. We believe that there can be a method to find the best clusters when reflecting the error cases that is able to guarantee the accuracy of task classifier and specialist networks together.

Also, based on our current result, three simple feedforward network can have a better performance compared to a convolutional neural network. It partly shows that using a series of neural network might be able to learn a task with much higher efficiently. We might try to a mathematical proof that adding more networks in CNNG will require much lower amount of training data compared to a deep neural network.

For the current implementation of CNNG with reflection, we only performed the reflection for once. We are interested in building a multi-layer CNNG, which will be a hierarchical group of neural networks. This could be another way for a deep neural network. It may be able to improve the performance for a further step.

\balance{} 
\bibliographystyle{SIGCHI-Reference-Format}
\bibliography{sample}

\end{document}